\documentclass[conference]{IEEEtran}
\IEEEoverridecommandlockouts
\usepackage{cite}
\usepackage{amsmath,amssymb,amsfonts}
\usepackage{graphicx}
\usepackage{textcomp}
\usepackage{xcolor}
\usepackage{xspace}
\usepackage{graphicx}
\usepackage{subcaption}
\usepackage{algorithm}
\usepackage{algpseudocode}
\usepackage{booktabs}

\def\BibTeX{{\rm B\kern-.05em{\sc i\kern-.025em b}\kern-.08em
    T\kern-.1667em\lower.7ex\hbox{E}\kern-.125emX}}

\newcommand{\model}{Loss-at-Risk\xspace}
\newcommand{\modelbf}{\textbf{Loss-at-Risk}\xspace}

\begin{document}

\title{Enhancing Risk Assessment in Transformers with Loss-at-Risk Functions}

\author{\IEEEauthorblockN{1\textsuperscript{st} Jinghan Zhang}
\IEEEauthorblockA{
\textit{Portland State University}\\
Portland, USA \\
jinghanz@pdx.edu}

\and
\IEEEauthorblockN{2\textsuperscript{nd} Henry Xie}
\IEEEauthorblockA{
\textit{Westview High School}\\
Portland, USA \\
henryjxie@gmail.com}

\and
\IEEEauthorblockN{3\textsuperscript{rd} Xinhao Zhang}
\IEEEauthorblockA{
\textit{Portland State University}\\
Portland, USA \\
xinhaoz@pdx.edu}

\and
\IEEEauthorblockN{4\textsuperscript{th} Kunpeng Liu}
\IEEEauthorblockA{
\textit{Portland State University}\\
Portland, USA \\
kunpeng@pdx.edu}
}

\maketitle

\begin{abstract}
In the financial field, precise risk assessment tools are essential for decision-making. Recent studies have challenged the notion that traditional network loss functions like Mean Square Error (MSE) are adequate, especially under extreme risk conditions that can lead to significant losses during market upheavals. Transformers and Transformer-based models are now widely used in financial forecasting according to their outstanding performance in time-series-related predictions. However, these models typically lack sensitivity to extreme risks and often underestimate great financial losses. To address this problem, we introduce a novel loss function, the Loss-at-Risk, which incorporates Value at Risk (VaR) and Conditional Value at Risk (CVaR) into Transformer models. This integration allows Transformer models to recognize potential extreme losses and further improves their capability to handle high-stakes financial decisions. Moreover, we conduct a series of experiments with highly volatile financial datasets to demonstrate that our Loss-at-Risk function improves the Transformers' risk prediction and management capabilities without compromising their decision-making accuracy or efficiency. The results demonstrate that integrating risk-aware metrics during training enhances the Transformers' risk assessment capabilities while preserving their core strengths in decision-making and reasoning across diverse scenarios.
\end{abstract}


\section{Introduction}

In the financial sector, decision support systems depend critically on precise and reliable risk assessments, especially in evaluating the maximum potential loss from a decision~\cite{wu2023bloomberggpt,li2023large,yang2023fingpt, di2024addressing,chang2022design}. Transformer models, particularly Large Language Models (LLMs) and extensive time series models based on Transformer architectures have emerged as key tools in complex risk management and decision analysis~\cite{10.1145/3687485,10415697,tian2024mmrec,gong2024neurosymbolicembeddingshorteffective,mao2023large}. These models typically utilize standard loss functions such as Mean Squared Error (MSE) for training and fine-tuning, which ensure high accuracy in predictions and judgments~\cite{chicco2021coefficient,jadon2022comprehensive,xing2024predicting,deng-etal-2022-title2event,yanenhancing,dingdata}. However, their loss functions in training and fine-tuning process are primarily designed to minimize average prediction errors, and they often overlook the potential for extreme risk events. This oversight can lead Transformers to make overly optimistic decisions during significant market upheavals or unusual situations, potentially resulting in substantial financial losses. Although extreme risk events are infrequent in the financial sector, they can cause severe economic losses and credit risks, such as financial crises and market collapse. Thus, it is essential to improve Transformers' awareness of risk and their risk assessment capabilities to better influence their decision-making processes.

Previous studies have addressed the issue of generalization for Transformer models handling small datasets by introducing prior knowledge and improved loss function methods~\cite{wang2020comprehensive,tian2022recent,mo2024surveyconversationalsearch,mo2024survey,mao2023learning}. For instance, some research enhances model generalization by quantifying uncertainty or using task-specific loss function variants to boost performance in particular scenarios~\cite{fu2024generalized,mo2024convgqrgenerativequeryreformulation,zhang2024dynamic,202410.1536,zhang2024tfwt}. However, these methods often fail to address the sensitivity of Transformer models to high-risk decisions directly. While these techniques can improve prediction accuracy under normal conditions, they also tend to overlook risk assessments under extreme conditions. This oversight can lead Transformers to make insufficiently cautious decisions when faced with potential risks.

These approaches primarily focus on improving models' handling of common data rather than optimizing responses to rare or extreme events~\cite{zhang2018generalized,zhang2024ratt,zhang2024thoughtspaceexplorernavigating}. This limitation is particularly critical in the financial sector, where even rare market anomalies can lead to major economic disruptions. Thus, traditional methods fall short in enabling Transformers to adequately assess and manage high-risk events that could result in severe consequences, affecting the safety and reliability of decisions at crucial moments.

Therefore, we consider that during the training process of Transformers, introducing risk-aware metrics would make the model more sensitive to potential extreme losses. For Transformers, the backpropagation is an optimal stage for such adjustments because the model has already acquired a general understanding of data patterns from its initial training on large datasets~\cite{kaplan2020scaling,cai2018exploring,howard2018universal,zhang2024dynamicadaptivefeaturegeneration,zhang2024tifgtextinformedfeaturegeneration}. During this process, incorporating specialized loss functions can refine the model's responses to be particularly cautious about rare but high-impact risks crucial in high-stakes environments like finance. This method leverages the model's ability to generalize from broad data while honing in on the nuanced understanding needed for specific tasks, which preserves the model's overall performance and enhances its precision in critical risk assessments.

\textbf{Our Target.} To address these problems, we target three main aspects: (1) We aim to make our Transformers more sensitive to extreme losses during the backpropagation stage. By integrating metrics into the loss function, we focus on minimizing losses in worst-case scenarios. This approach is crucial for high-stakes applications where errors can lead to significant consequences. (2) We need to ensure these changes do not compromise the Transformer's core strengths. This includes its time-series-related prediction, attention mechanism and robustness. The model must still excel across standard tasks while managing the demands of high-risk environments. We manage the backpropagation process to boost the Transformer's overall abilities. (3) We then rigorously test the updated loss function in financial settings. We evaluate the model performance on financial datasets marked by high volatility and risk, to compare how the model performs before and after introducing the new loss functions. Our goal is to confirm that our changes enhance the model’s ability to predict and manage risks without losing accuracy or efficiency in decision-making.

\begin{figure}[h]
    \centering
    \includegraphics[width=0.9\linewidth]{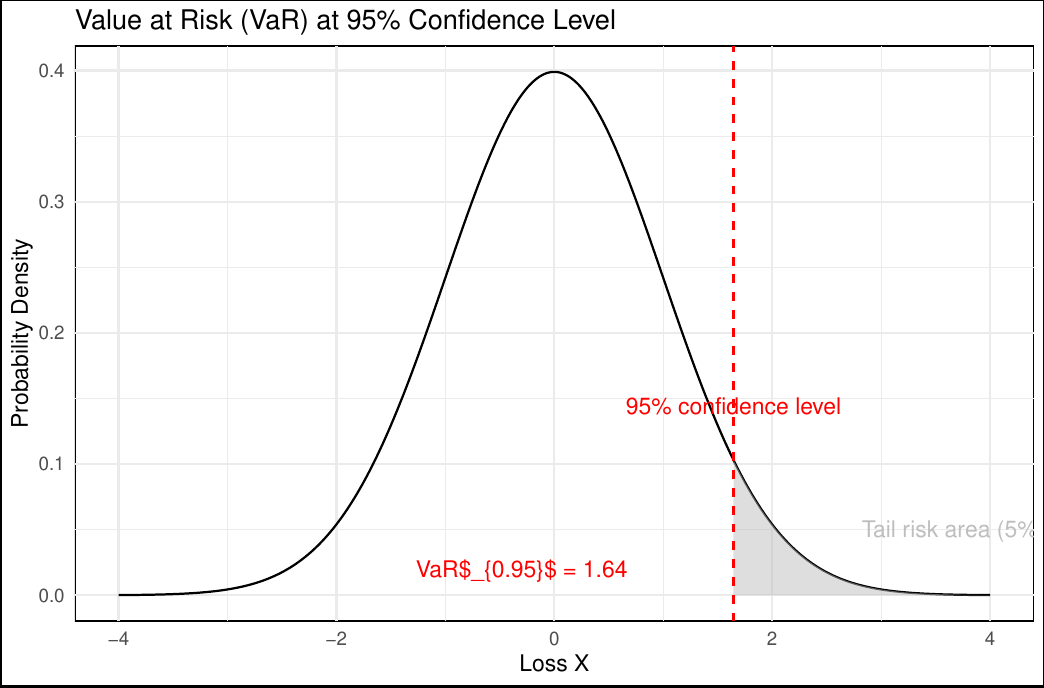}
    \caption{A schematic diagram of VaR.}
    \label{fig:VaR}
\end{figure}

\textbf{Our Method.} According to the three targets, we design \modelbf function specifically to enhance the robustness of Transformers in high-stakes financial scenarios. Our method integrates both Value at Risk (VaR)~\cite{benninga1998value,better2008simulation,ismal2010volatility} and Conditional Value at Risk (CVaR)~\cite{tamar2015optimizing,chow2015risk,alexander2004comparison} into the traditional MSE loss framework. (1) We enhance sensitivity to extreme losses. The \model calculates the traditional MSE for average scenarios and adds a weighted component based on VaR and CVaR metrics. VaR provides a threshold beyond which a given percentage of losses will fall, introducing a focus on worst-case scenarios. CVaR goes further by averaging the losses that exceed the VaR threshold, ensuring the model accounts for the severity of extreme outcomes. By focusing on these metrics during backpropagation, the model becomes more sensitive to major potential losses, while ignoring common, less critical errors. (2) We also maintain core capabilities at the Transformer's loss function. While integrating these risk metrics, we carefully calibrate their influence to maintain the Transformer's essential skills in reasoning, decision-making, and robustness across general tasks. The weighting of the VaR and CVaR components within the \model function is adjusted to ensure that risk management enhancements do not compromise the model’s effectiveness in less volatile environments. This balanced approach allows the model to remain versatile and effective, operating reliably in typical and high-risk financial contexts. (3) Furthermore, to validate the effectiveness of our \model function, we implement a rigorous testing regime using diverse financial datasets. By comparing the model's performance on these datasets before and after applying the Loss-at-Risk function, we assess how well the model predicts and manages financial risks. The goal is to demonstrate that our approach not only preserves but enhances the model's decision-making accuracy and efficiency under stress, and further proving that the model can handle the complexities of real-world financial applications effectively.

In summary, our contribution includes:

\begin{enumerate}
    \item We integrate risk assessment with MSE loss for Transformer models' training process to enhance their ability in extreme loss scenarios. This \model equips Transformers the risk-awareness to handle high-stakes financial risks effectively.
    
    \item Our method maintains the Transformer model's core decision-making and time-series-based predicting capabilities while enhancing risk sensitivity. This ensures effective performance in both typical and high-risk financial situations.
    
    \item We conduct rigorous testing on volatile financial datasets to validate our model function. The results confirm improved risk management without sacrificing decision-making accuracy or efficiency.
\end{enumerate}

\section{Related Work}
\subsection{Transformers in Time Series Data}

Transformers have revolutionized the analysis of time series data due to their ability to model complex dependencies across time steps. Their architecture, particularly effective in capturing long-range interactions in data sequences, has been extensively applied in various domains, including financial markets~\cite{vaswani2017attention,brown2020language,zhang2024prototypical,yan2024machine,yan2019predict,zhang2024tifg}. However, Transformers require substantial data to train effectively, which poses a challenge in scenarios where time series data is limited or highly volatile, such as in financial forecasting during unusual market conditions. The primary limitation of Transformers in these contexts is their potential overfitting to small datasets and their sensitivity to the noise commonly found in financial time series. Traditional training methods often fail to equip Transformers with the robustness needed to handle the unpredictability and sparse data points characteristic of financial time series~\cite{CHEN20153142,CHEN2017340,dan2024evaluation,CHOWDHURY2023109314,wang2004improving,yeung2002improving}.


\subsection{Loss Function for Transformers}
In research on Transformers, the predominant loss functions typically include classic forms such as cross-entropy~\cite{zhang2018generalized,wang2024machine,feng2021can} and MSE~\cite{wang2009mean,kim2021comparing,song2023going,wang2024research}. These are often augmented with task-specific variants to tackle challenges like data imbalance\cite{mustafa2022training,kang2022tie,yang2021task} or to facilitate multi-task learning~\cite{wang2016learning,gao2016novel}. Our work marks a significant departure by incorporating risk assessment into the loss function during the backpropagation process, specifically enhancing the model's ability to manage risks in scenarios where the stakes are exceptionally high. Integrating risk assessment with traditional loss functions can quantify and minimize potential extreme losses. Moreover, with this integrated loss, we introduce a novel dimension to Transformer training that prioritizes risk sensitivity—an essential factor for applications in fields like finance, where accurately predicting and mitigating risks can have profound implications.

\subsection{Value at Risk and Conditional Value at Risk}
VaR is a statistical measure used to assess the level of financial risk within a firm or investment portfolio over a specific time frame, as we show in Fig.~\ref{fig:VaR}. It is defined as the maximum loss expected (under normal market conditions) over a target horizon within a given confidence interval. Mathematically, VaR is represented as:
\begin{equation}
\text{VaR}_{\alpha}(X) = \inf \left\{ x \in \mathbb{R} : P(X > x) \leq 1 - \alpha \right\},
\end{equation}
where $ X $ represents the loss random variable and $ \alpha $ is the confidence level. VaR is widely used due to its ability to provide a clear metric that quantifies potential losses at a specific confidence level.

CVaR, also known as Expected Shortfall (ES), measures the expected loss assuming that a loss is greater than the VaR. It is considered a more coherent and risk-sensitive measure than VaR because it accounts for the tail risk beyond the VaR threshold. CVaR is defined as the average of losses that exceed the VaR value at a certain confidence level:
\begin{equation}
\text{CVaR}_{\alpha}(X) = E[X | X > \text{VaR}_{\alpha}(X)].
\end{equation}
Alternatively, CVaR can be calculated through the integration of the tail distribution of losses:
\begin{equation}
\text{CVaR}_{\alpha}(X) = \frac{1}{1 - \alpha} \int_{\alpha}^{1} \text{VaR}_{u}(X) \, du,
\end{equation}
where $ u $ spans from the confidence level $ \alpha $ to 1.

\begin{figure}
    \centering
    \includegraphics[width=0.9\linewidth]{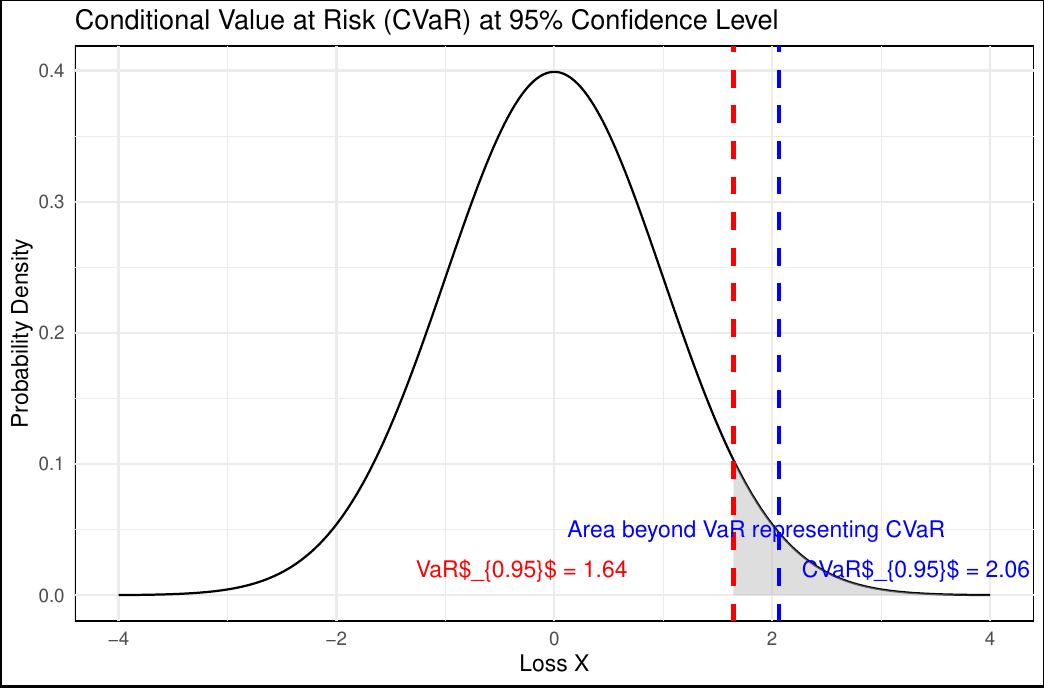}
    \caption{A schematic diagram of CVaR.}
    \label{fig:VaR}
\end{figure}

\section{Methodology}
\subsection{Methodology Overview}
In this section, we break down our approach, focusing on how we designed and tested a new loss function for LLMs, combining MSE with VaR. We keep things practical by detailing the key math concepts and their proofs, pinpointing exactly how this helps in training LLMs. We also introduce CVaR, explaining how it builds on VaR to cover the average of more extreme losses. This part also tackles the specific challenges we faced in optimizing the model and how we solved them.
\subsection{Loss Function Definition}
\paragraph{\model with VaR}
We define our primary loss function $ L(y, y_{\text{true}}) $ using MSE, given by:
\begin{equation}
\text{MSE}(y, y_{\text{true}}) = (y - y_{\text{true}})^2,
\label{Eq:MSE}
\end{equation}
where $ y $ is the model's output and $ y_{\text{true}} $ is the actual value. MSE is favored in regression tasks due to its emphasis on penalizing larger errors more significantly, which helps in stabilizing the training process.

For risk-sensitive applications, we incorporate the concept of VaR, defined at a confidence level $ \alpha $. VaR measures the maximum loss not exceeded with a probability of $ \alpha $, thus focusing on worst-case scenarios:
\begin{equation}
\text{VaR}_{\alpha}(L) = \inf\{\xi \in \mathbb{R} : P(L(y, y_{\text{true}}) \leq \xi) \geq \alpha\}.
\end{equation}

To combine these measures into a single coherent loss function suitable for LLMs, we introduce the \model:
\begin{equation}
L_{\text{VaR-MSE}}(y, y_{\text{true}}) = L_{\text{MSE}}(y, y_{\text{true}}) + \lambda \cdot VaR_{\alpha}(L_{\text{MSE}}),
\end{equation}
where:
\begin{itemize}
    \item $L_{\text{VaR-MSE}}(y, y_{true})$ is the combined MSE and VaR loss function.
    \item $\lambda$ is a parameter that balances the impact of the MSE and VaR components.
    \item $VaR_{\alpha}(L_{\text{MSE}})$ measures the maximum loss at the confidence level $\alpha$, based on the MSE calculation.
\end{itemize}
This hybrid function is designed to optimize standard performance and reduce the risk of extreme losses, allowing for adjustments based on the specific risk management needs of the application.

\paragraph{\model with CVaR}

Building on our initial \model with VaR framework, we further enhance it by incorporating CVaR, defined at a confidence level $ \alpha $. CVaR is advantageous as it calculates the mean loss beyond the VaR threshold, offering a more comprehensive measure of tail risks:
\begin{equation}
\text{CVaR}_{\alpha}(L) = \frac{1}{1-\alpha} \int_{\alpha}^{1} \text{VaR}_{u}(L) \, du,
\end{equation}
which captures the average of the worst $ 100(1-\alpha)\% $ outcomes, making it a stringent measure for scenarios with potentially catastrophic financial implications.

To tailor our loss function for heightened risk sensitivity in LLMs, we propose the CVaR-enhanced Loss-at-Risk function:
\begin{equation}
L_{CVaR-MSE}(y, y_{\text{true}}) = L_{MSE}(y, y_{\text{true}}) + \lambda \cdot CVaR_{\alpha}(L_{MSE}),
\end{equation}
where $ \lambda $ finely tunes the balance between MSE and CVaR, facilitating a targeted approach to manage extreme risk exposures effectively.

This adaptation of the loss function allows LLMs to remain robust in their predictive capabilities while enhancing their precision in assessing and mitigating extreme risks. CVaR's focus on the average of the most severe losses provides a more realistic and useful risk assessment, particularly suitable for financial sectors and other high-stake applications where overlooking such risks could lead to significant repercussions.

\subsection{Mathematical Rigor and Suitability}

\paragraph{Continuity and Differentiability}
In \model, the MSE part is a quadratic function, and all polynomial functions are continuous and differentiable across their entire domain. This property is crucial for the application of gradient-based optimization methods. The derivative of MSE is given by:
\begin{equation}
\frac{d}{dy}L_{\text{MSE}}(y, y_{\text{true}}) = 2(y - y_{\text{true}}),
\end{equation}
demonstrating smoothness and suitability for such methods.

The continuity of the VaR component depends on the continuity of the loss distribution. If the loss distribution is continuous, the Cumulative Distribution Function (CDF) is also continuous, and hence VaR is continuous.

Differentiability of VaR can be complex and is contingent on the continuity and non-zero property of the Probability Density Function (PDF) at the quantile. Assuming a smooth distribution, VaR can be approximated as differentiable.

For CVaR, the continuity and differentiability are similarly conditioned on the underlying distribution of losses. Since CVaR is essentially an average of VaR over a range of confidence levels above $ \alpha $, its continuity depends directly on the continuity of VaR across these levels. This typically means that if VaR is continuous for a given distribution of losses, CVaR will also be continuous.

The differentiability of CVaR, however, can be more nuanced due to its dependence on the integral of quantiles which themselves may have points of non-differentiability depending on the loss distribution's characteristics. Nevertheless, assuming that the loss distribution is smooth and well-behaved, CVaR can often be treated as differentiable for practical purposes in optimizing models' risk awareness. This allows the use of gradient-based methods in training, optimizing both the average and the extreme risks associated with predictions.

\paragraph{Convexity} The convexity of the MSE is validated by its second derivative:
\begin{equation}
\frac{d^2}{dy^2} L_{\text{MSE}}(y, y_{\text{true}}) = 2,
\end{equation}
which is positive, confirming that MSE is convex.

The convexity of VaR depends on the characteristics of the loss distribution. Under regular conditions (e.g., unimodal, not heavily skewed distributions), VaR is typically convex. Thus in \model, the function's continuity and differentiability are typically preserved, with convexity contingent on the convexity of the VaR component.

The convexity of CVaR, much like VaR, is contingent upon the characteristics of the underlying loss distribution. CVaR, defined as the expected value of the losses exceeding the VaR threshold, also tends to be convex under conditions where the loss distribution is smooth and regular. This integration over a range of VaR values tends to smooth out irregularities, potentially enhancing the convexity of CVaR compared to VaR alone. Therefore, when the loss distribution is unimodal and not heavily skewed, the CVaR component generally exhibits convexity. Consequently, when combined with MSE in the Loss-at-Risk function, the overall function maintains desirable convex properties, assuming the parameter $\lambda$ is chosen appropriately. In this way, we confirm that the \model function is mathematically robust and suitable for use in gradient-based optimization scenarios.

\paragraph{Robustness and Sensitivity}

(1) Impact on accuracy. The Mean Squared Error, $L_{\text{MSE}}(y, y_{\text{true}})$, ensures that the model is penalized for any deviation from the true values, which inherently promotes high accuracy in predictions. The key property here is that MSE treats all errors equally, squaring the magnitude of deviations, which emphasizes larger errors more heavily, thus preventing large deviations in model predictions. (2) Enhanced sensitivity to extreme risks. The addition of the VaR or CVaR component introduces a focus on the tail of the loss distribution. Specifically, VaR focuses on the maximum potential loss that can occur with a certain probability, while CVaR provides an average of the worst losses beyond the VaR threshold. This aspect of the Loss-at-Risk function causes the model to not only minimize average prediction errors but also to be particularly cautious about the potential for significant outliers or extreme values in predictions.

\subsection{Optimization in Transformers}

Then we explore the mathematical underpinnings of the MSE and VaR in the context of Transformers. Our aim is to analyze the convexity of VaR concerning MSE and outline the optimization conditions for this model.

We start by establishing the risk measure function $ V(\xi) $ to understand VaR's role in Transformers:
\begin{equation}
V(\xi) = E\left[\xi + \frac{1}{1 - \alpha}(L(X) - \xi)_+\right],
\end{equation}
where $ (z)_+ $ represents the non-negative part, expressed as $ \max(z, 0) $.

\paragraph{Derivative of VaR} The derivative $ V'(\xi) $ is crucial for understanding VaR's behavior in the optimization process:
\begin{equation}
V'(\xi) = \frac{d}{d\xi}E\left[\xi + \frac{1}{1 - \alpha}(L(X) - \xi)_+\right],
\end{equation}
The function $ (L(X) - \xi)_+ $ is piecewise linear, with its derivative breaking down:
\begin{itemize}
  \item When $ L(X) > \xi $, $ (L(X) - \xi)_+ = L(X) - \xi $ and its derivative with respect to $ \xi $ is -1.
  \item When $ L(X) \leq \xi $, $ (L(X) - \xi)_+ = 0 $, derivative is 0.
\end{itemize}
Thus, we derive:
\begin{equation}
V'(\xi) = 1 - \frac{1}{1 - \alpha} P(L(X) > \xi).
\end{equation}

\paragraph{Optimization Condition} The condition for optimization $ V'(\xi) = 0 $ is essential for setting the correct threshold in VaR:
\begin{equation}
1 - \frac{1}{1 - \alpha} P(L(X) > \xi) = 0.
\end{equation}
Reformulating gives:
\begin{equation}
P(L(X) > \xi) = 1 - \alpha.
\end{equation}
This indicates that we need to find a specific $ \xi $, where the probability of the model's loss exceeding $ \xi $ matches $ 1 - \alpha $. This threshold is crucial in LLM optimization as it delineates the boundary beyond which the model's loss is unlikely to exceed under the chosen confidence level $ \alpha $.

This rigorous approach ensures that the Loss-at-Risk function, when calibrated with the appropriate $ \xi $ value, effectively balances between minimizing typical prediction errors and preventing large unexpected losses, hence optimizing the LLM's performance in risk-sensitive applications.

\begin{algorithm}
\caption{\model}
\begin{algorithmic}[1]
\State \textbf{Input:} Predictions $y_{\text{pred}}$, True values $y_{\text{true}}$, Risk level $\alpha$, Weight $\lambda$
\State \textbf{Output:} Loss-at-Risk $L_{VaR-MSE}(y, y_{\text{true}})$

\State Initialize MSE loss  $L_{\text{MSE}} \leftarrow \emptyset$

\For{each sample $i_{\text{pred}}$,$i_{\text{true}}$ in $y_{\text{pred}}$ and $y_{\text{true}}$}

\State Compute ${\text{MSE}}(i_{\text{pred}}, i_{\text{true}})$ according to Eq.\ref{Eq:MSE}

\State $L_{\text{MSE}} \leftarrow {\text{MSE}}(i_{\text{pred}}, i_{\text{true}})$

\EndFor

\State Calculate the $\alpha$-quantile of $L_{\text{MSE}}$: 
\Statex $L_{\text{VaR}} \leftarrow \text{quantile}(L_{\text{np}}, \alpha)$

\State Compute the mean of $L_{\text{MSE}}$: $\overline{L_{\text{MSE}}} \leftarrow \text{mean}(L_{\text{MSE}})$

\State Compute the total loss with VaR weighting: 
\Statex $L_{\text{VaR-MSE}}(y, y_{\text{true}}) \leftarrow \overline{L_{\text{MSE}}} + \lambda \cdot L_{\text{VaR}}$
\State \textbf{return} $L_{\text{VaR-MSE}}(y, y_{\text{true}})$

\end{algorithmic}
\end{algorithm}




\section{Experiments}

In the experimental section of our study, we conduct a series of tests to evaluate the performance of the Loss-at-Risk function in various scenarios. We aim to answer three key questions through the following experiments:
\begin{enumerate}
    \item How does the Loss-at-Risk function compare in baseline performance to traditional MSE loss on standard tasks?
    \item How effective is the \model function in handling extreme values?
    \item What is the impact of key parameters such as the confidence level $\alpha$ and weight $\lambda$ in the \model function on the performance of the models?
\end{enumerate}
In answering these questions, we design the following experiments: 1) Baseline Performance Comparison. We use the standard financial dataset FNSPID~\cite{dong2024fnspid} from Nasdaq, train models using both MSE loss and the Loss-at-Risk function and compare their accuracy. 2) Performance Under Extreme Conditions. We perform the models on financial transaction data during extreme market conditions, fine-tune the models using both MSE and Loss-at-Risk, and compare their performance on accuracy and sensitivity to high-risk scenarios. 3) Parameter Sensitivity Analysis. We vary $\alpha$ and $\lambda$ while keeping other conditions fixed during fine-tuning, assess their performance under different parameter settings, and illustrate the impact of parameter changes on model performance through graphs.

\subsection{Experimental Settings}

\paragraph{Datasets} For the first experiment, we use the FNSPID dataset in our experiments, which combines stock prices and financial news across a broad range of S\&P 500 companies. This dataset features over 29.7 million stock price records and 15.7 million news entries from 1999 to 2023. Specifically, we focus on the stocks with tickers KO, AMD, TSM and WMT. These stocks cover various sectors, from beverages to big tech, providing a diverse view of market reactions to news. We choose this dataset to explore how news impacts stock behavior and to leveraging its rich time-series data to analyze the direct effects of financial news on stock movements.

\paragraph{Baseline Performance Comparison} 
\begin{figure*}[h]
    \centering
    \begin{subfigure}[b]{0.25\linewidth}
        \centering
        \includegraphics[width=\linewidth]{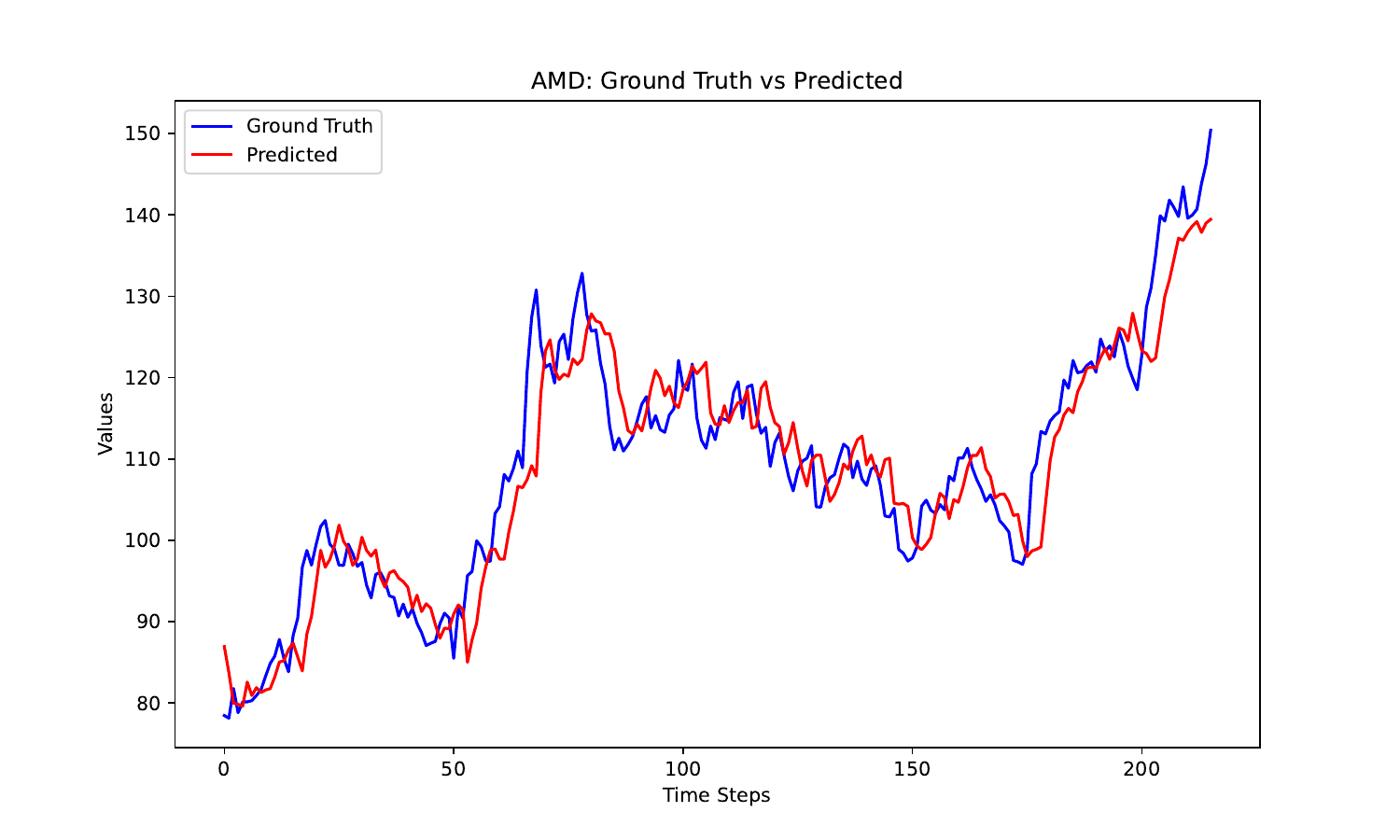}
        \caption{AMD with MSE loss}
        \label{fig:AMD_MSE}
    \end{subfigure}%
    \begin{subfigure}[b]{0.25\linewidth}
        \centering
        \includegraphics[width=\linewidth]{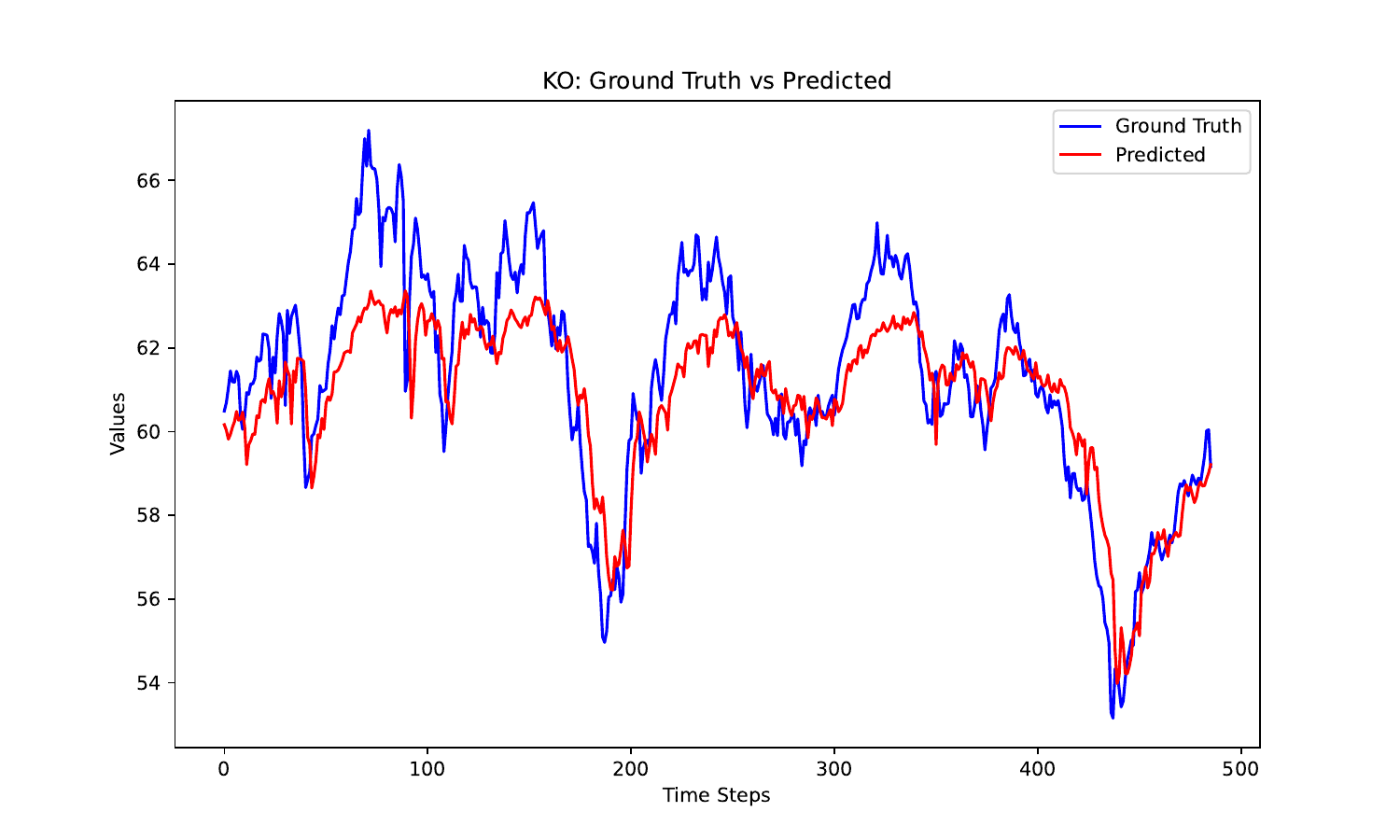}
        \caption{KO with MSE loss}
        \label{fig:KO_MSE}
    \end{subfigure}%
    \begin{subfigure}[b]{0.25\linewidth}
        \centering
        \includegraphics[width=\linewidth]{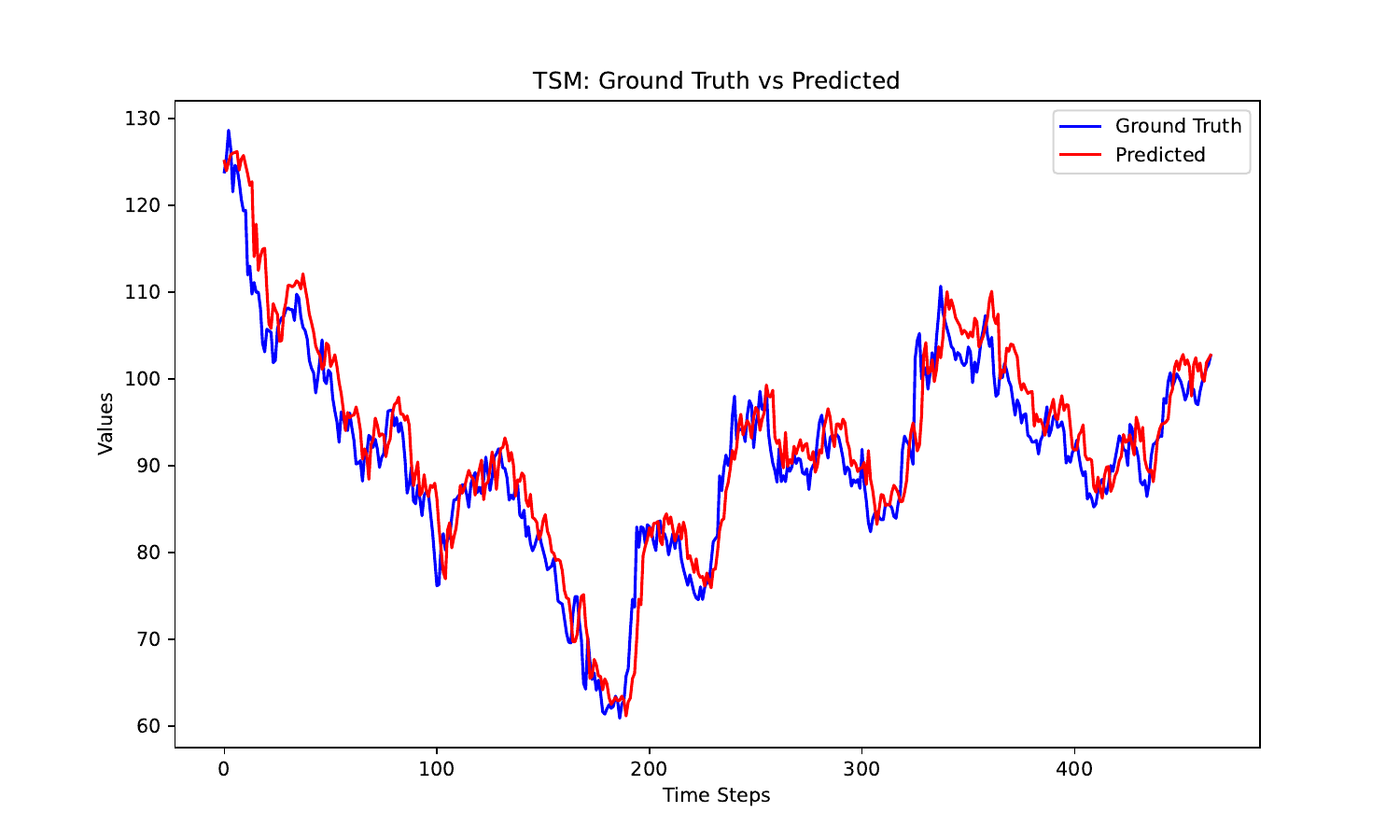}
        \caption{TSM with MSE loss}
        \label{fig:TSM_MSE}
    \end{subfigure}%
    \begin{subfigure}[b]{0.25\linewidth}
        \centering
        \includegraphics[width=\linewidth]{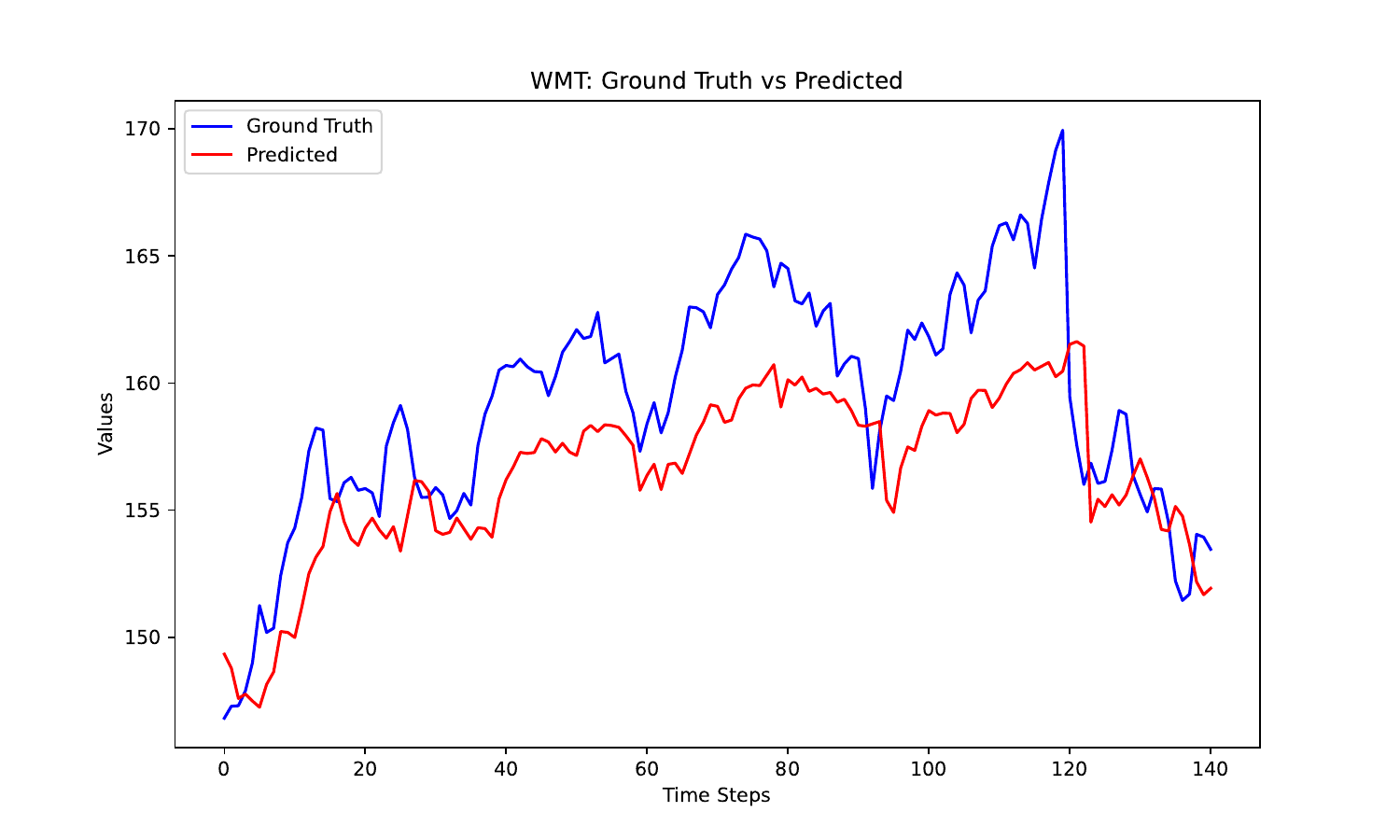}
        \caption{WMT with MSE loss}
        \label{fig:WMT_MSE}
    \end{subfigure}

    \begin{subfigure}[b]{0.25\linewidth}
        \centering
        \includegraphics[width=\linewidth]{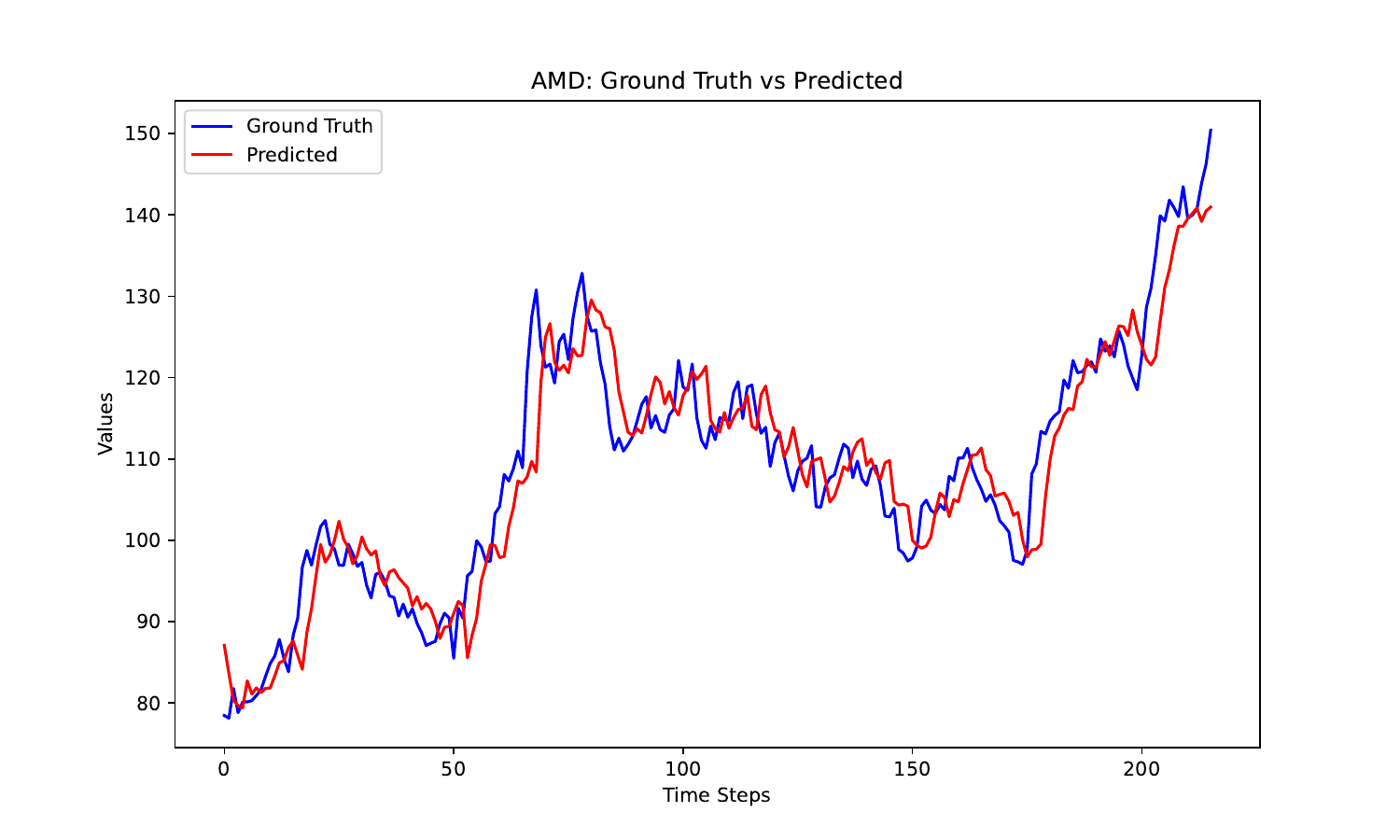}
        \caption{AMD with \model loss}
        \label{fig:AMD_VAR}
    \end{subfigure}%
    \begin{subfigure}[b]{0.25\linewidth}
        \centering
        \includegraphics[width=\linewidth]{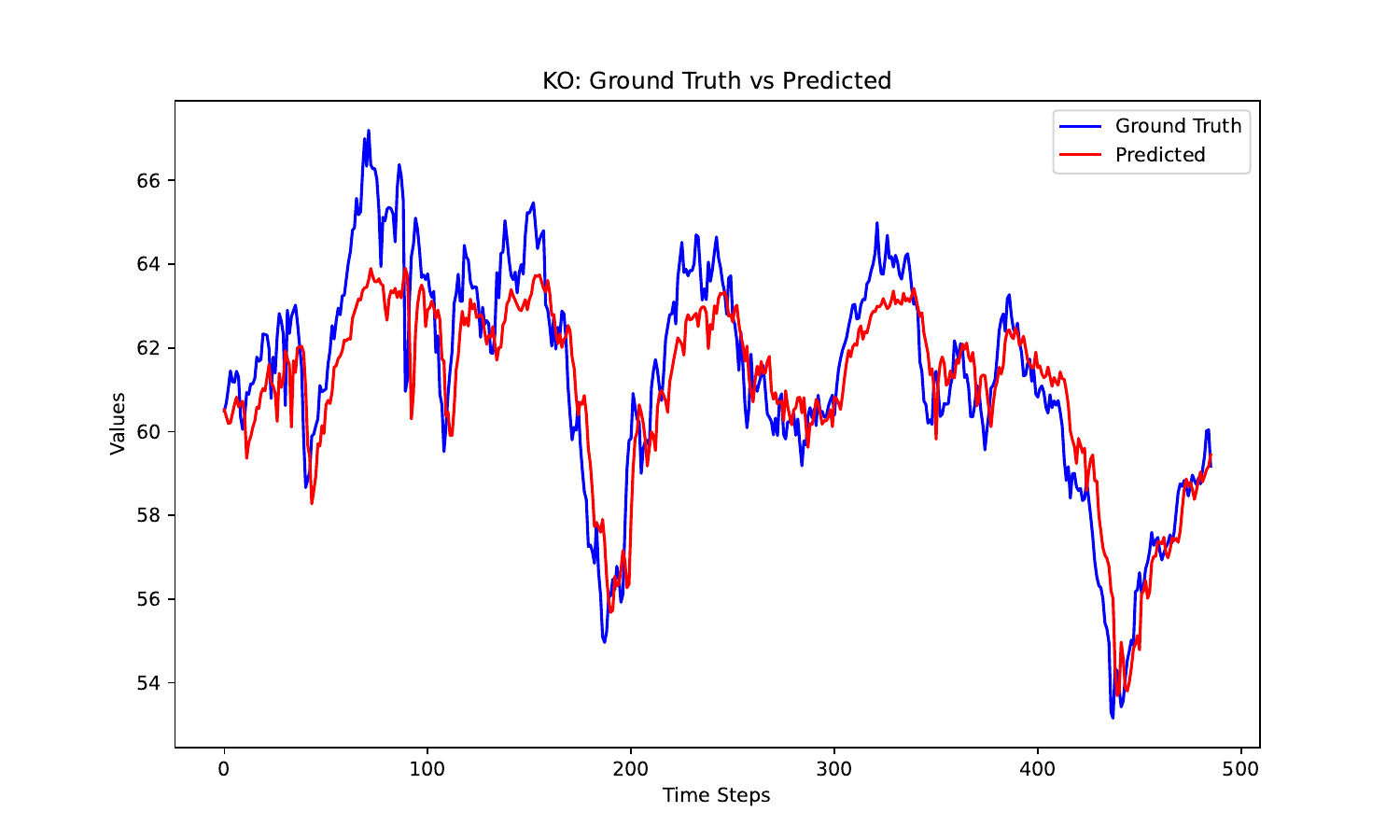}
        \caption{KO with \model loss}
        \label{fig:KO_VAR}
    \end{subfigure}%
    \begin{subfigure}[b]{0.25\linewidth}
        \centering
        \includegraphics[width=\linewidth]{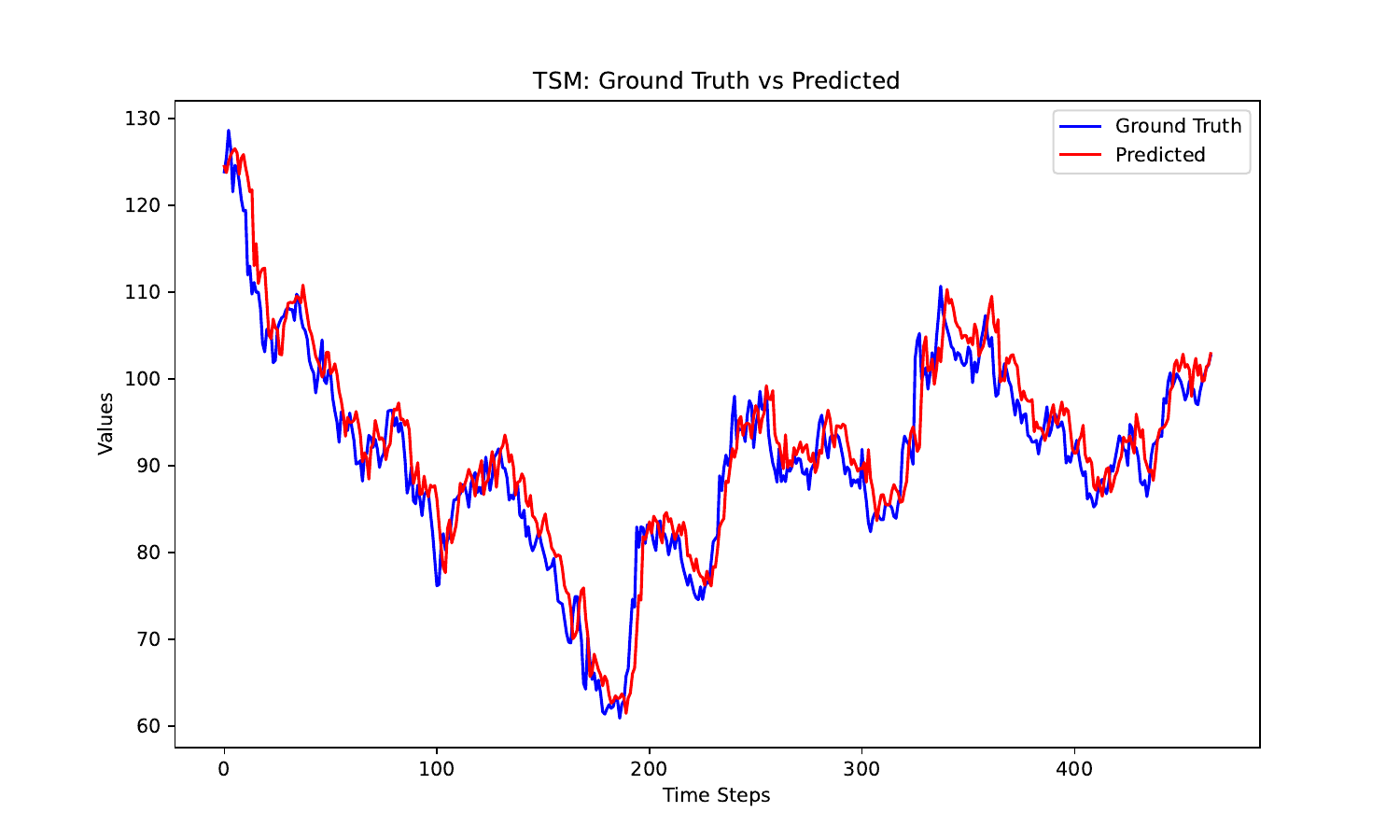}
        \caption{TSM with \model loss}
        \label{fig:TSM_VAR}
    \end{subfigure}%
    \begin{subfigure}[b]{0.25\linewidth}
        \centering
        \includegraphics[width=\linewidth]{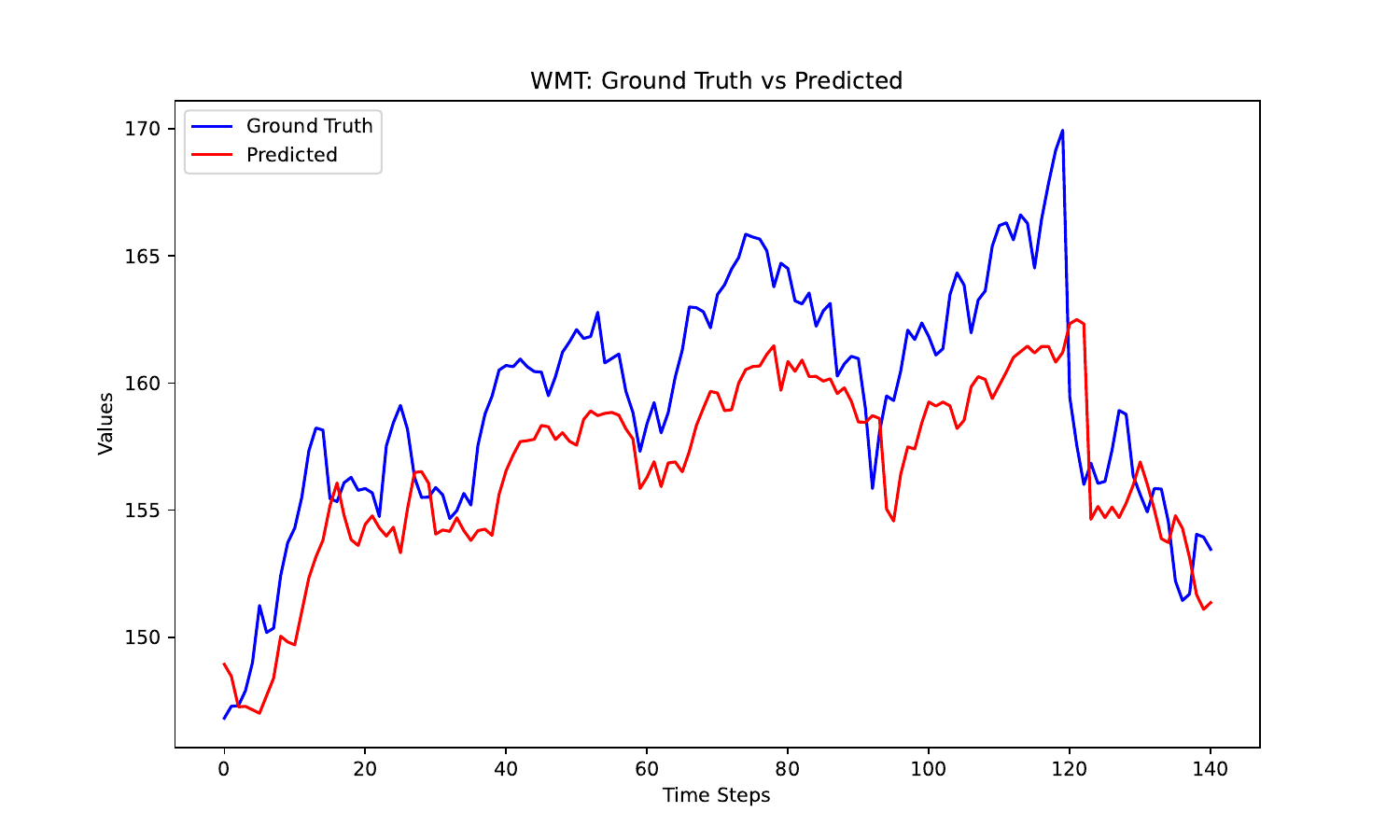}
        \caption{WMT with \model loss}
        \label{fig:WMT_VAR}
    \end{subfigure}

    \caption{Comparison of MSE and Loss-at-Risk performance across selected stocks on Transformer.}
    \label{fig:stocks_comparison}
\end{figure*}
\begin{table}[h]
    \centering
    \caption{Analysis of AMD's overall and extreme performance (top and bottom 5\% data). The best results are highlighted in \textbf{bold}, and the runner-up results are highlighted in \underline{underline}.}
    \renewcommand{\arraystretch}{1.3}
    \begin{tabular}{lccccc}
    \toprule
    \textbf{Model} & \textbf{MSE} & \textbf{MAE} & \textbf{R2} & \textbf{Max AE} & \textbf{Min AE} \\
    \midrule
    $L_{\text{MSE}}$ & 20.1314 & 3.2918 & 0.9790 & 8.5783 & \underline{5.2252} \\
    $L_{\text{VaR-MSE}}$ & \underline{19.0661} & \underline{3.2640} & \underline{0.9801} & \textbf{7.9585} & 5.7073 \\
    $L_{\text{CVaR-MSE}}$ & \textbf{18.8994} & \textbf{3.1816} & \textbf{0.9803} & \underline{8.0151} & \textbf{4.6462} \\
    \bottomrule
    \end{tabular}
    \label{tab:AMD}
\end{table}

We conduct an analysis on the AMD stock, specifically focusing on the highest and lowest $5\%$ of data points, corresponding to an $\alpha$ value of $5\%$, to simulate the performance under extreme conditions. The metrics we evaluated include MSE-VaR loss and MSE-CVaR loss. The outcomes of this analysis are presented in Table~\ref{tab:AMD} to show how these models perform under stress scenarios.

The Max Absolute Error (Max AE) and Min Absolute Error (Min AE) are metrics designed to capture the absolute discrepancies between predicted values $ y_{\text{pred}}(t) $ and true values $ y_{\text{true}}(t) $ at specific points in time. Specifically, Max AE is the absolute error computed at the time $ t $ when $ y_{\text{true}} $ reaches its maximum value, while Min AE is calculated at the time $ t $ when $ y_{\text{true}} $ reaches its minimum. 

\begin{equation}
\begin{aligned}
t_{\text{max}} &= \arg\max_t y_{\text{true}}(t), \\
\text{Max AE} &= \left| y_{\text{true}}(t_{\text{max}}) - y_{\text{pred}}(t_{\text{max}}) \right|.
\end{aligned}
\end{equation}

\begin{equation}
\begin{aligned}
t_{\text{min}} &= \arg\min_t y_{\text{true}}(t), \\
\text{Min AE} &= \left| y_{\text{true}}(t_{\text{min}}) - y_{\text{pred}}(t_{\text{min}}) \right|.
\end{aligned}
\end{equation}

\subsection{Standard Prediction}
In this experiment, our goal is to compare the performance of Transformers using the Loss-at-Risk function with those using traditional MSE loss regarding prediction accuracy. This is to evaluate whether the new loss function impacts the model's overall performance. Here we show that introducing sensitivity to extreme risks does not compromise the model's performance on routine tasks. As shown in Fig.~\ref{fig:stocks_comparison}, Transformers with Loss-at-Risk exhibit similar or slightly better performance than those using MSE loss. These results indicate that the Loss-at-Risk function enhances the model's ability to perceive extreme risks without degrading its overall predictive capabilities. These findings support applying this new loss function in risk-sensitive financial decision-making models.
\subsection{Performance on Extreme Values}
Then we test and evaluate the performance of Transformer with Loss-at-Risk under extreme value conditions and in predicting potential risk points. Both Loss-at-Risk and traditional MSE loss functions are evaluated on same Transformer and we observe their predictive behaviors during high-volatility scenarios.

\begin{figure}[htbp]
    \centering
    \vspace{-5mm}
    \includegraphics[width=0.9\linewidth]{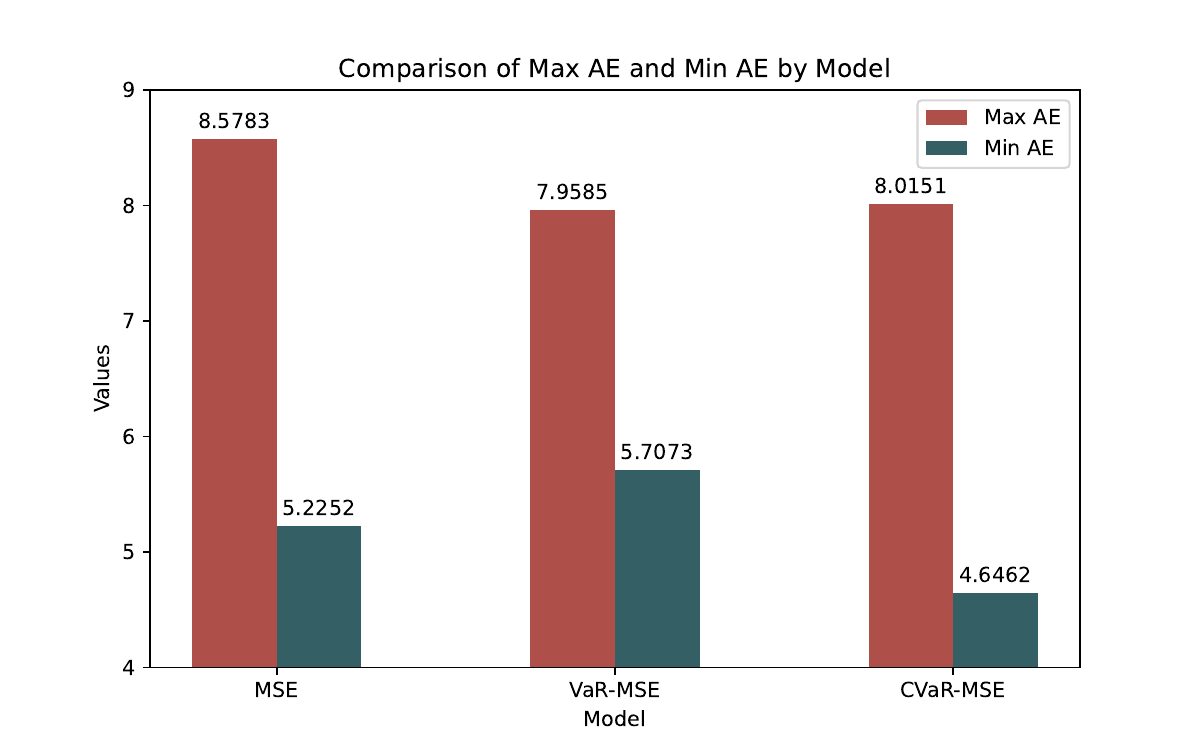}
    \caption{Model performance under extreme values (top and bottom $5\%$). Lower values indicate better outcomes.}
    \label{fig:model_performance_extreme}
    \vspace{-5mm}
\end{figure}

\begin{figure*}[htbp]
    \centering
    \includegraphics[width=0.98\linewidth]{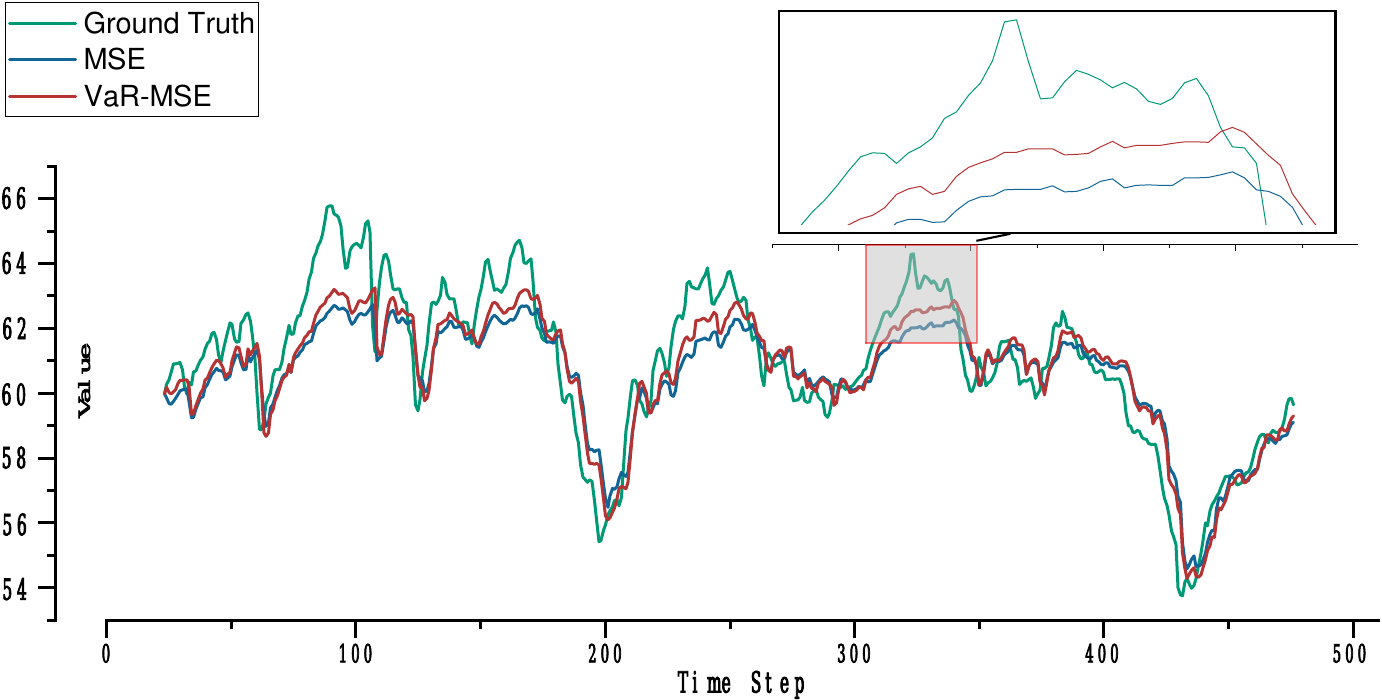}
    \caption{Model performance under extreme values (top and bottom $5\%$). Lower values indicate better outcomes.}
    \label{fig:scale}
    \vspace{-4mm}
\end{figure*}

As shown in Fig.~\ref{fig:model_performance_extreme} and Fig.~\ref{fig:scale}, we observe a notable improvement in model performance with Loss-at-Risk than MSE. Specifically, the results demonstrate that the maximum absolute error (Max AE) and minimum absolute error (Min AE) are lower for models utilizing the Loss-at-Risk function, particularly with the CVaR-MSE configuration. The Max AE for the VaR-MSE model showed a significant decrease compared to the standard MSE model, indicating a better handling of worst-case scenarios. Similarly, the CVaR-MSE model demonstrated the lowest Min AE, which suggests superior performance in managing the most favorable loss outcomes under extreme conditions.

\begin{figure}[htbp]
    \centering
    \begin{subfigure}[b]{\columnwidth}  
        \includegraphics[width=\textwidth]{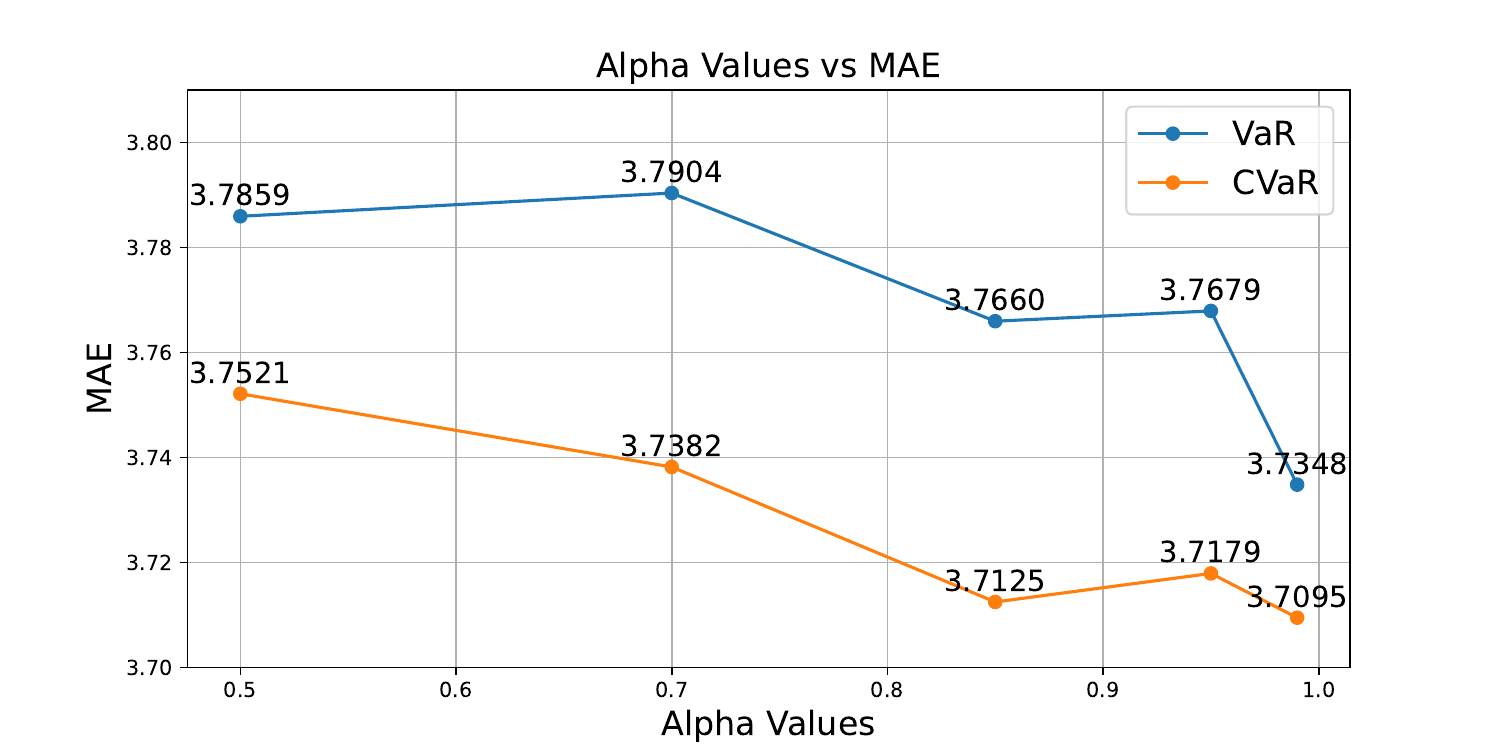}
        \caption{Variations in $\alpha$.}
        \label{fig:lambda_var_mse}
    \end{subfigure}
    \begin{subfigure}[b]{\columnwidth}  
        \includegraphics[width=\textwidth]{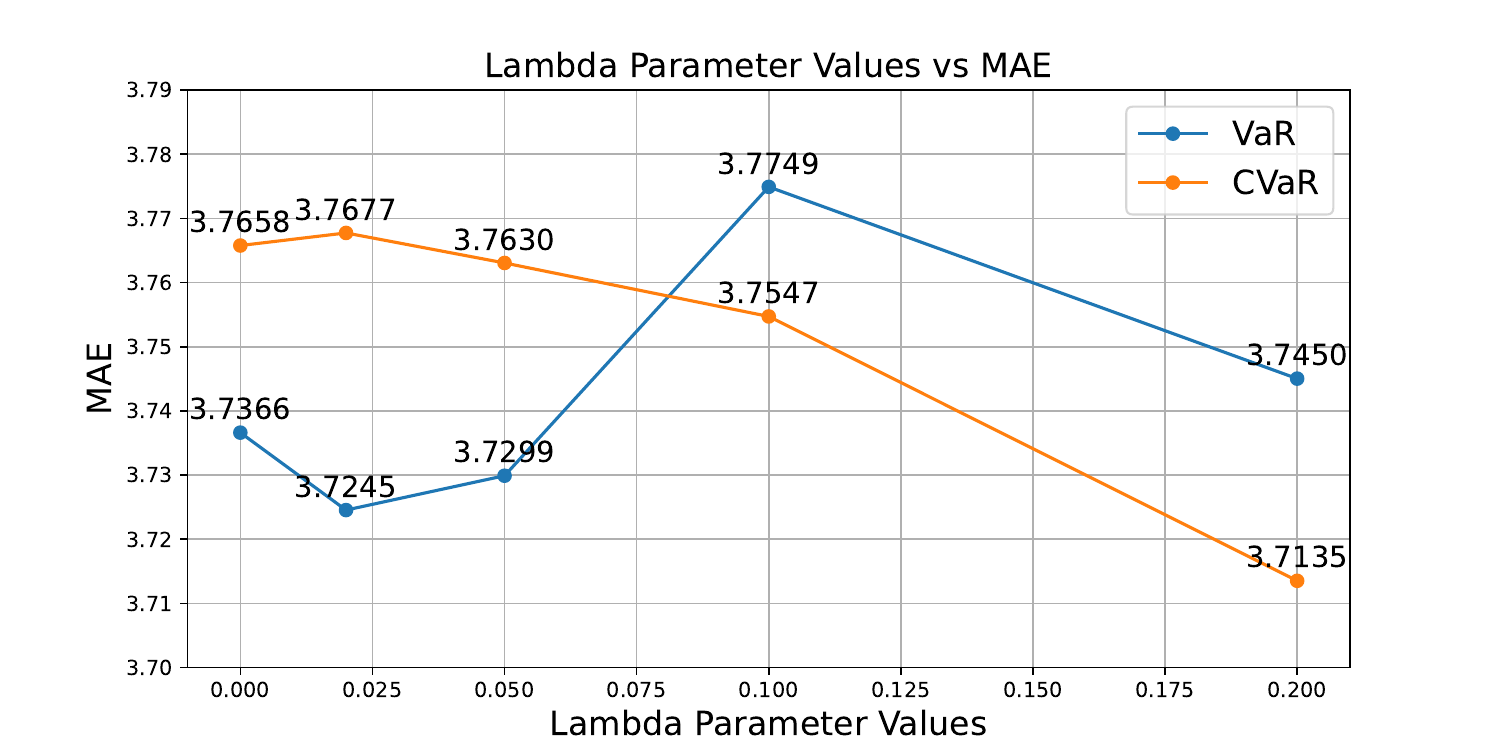}
        \caption{Variations in $\lambda$.}
        \label{fig:alpha_cvar_mse}
    \end{subfigure}
    \caption{$L_{\text{VaR-MSE}}$ and $L_{\text{CVaR-MSE}}$ performance on the AMD dataset for as $\lambda$ and $\alpha$ variations.}
    \label{fig:both_mse}
    \vspace{-4mm}
\end{figure}

\subsection{Ablation Study}
In the ablation study, we examine the impact of varying $\alpha$ values on the performance of Loss-at-Risk functions that incorporate VaR and CVaR. The $\alpha$,  confidence level for quantifying risk, is crucial in determining how the model perceives and manages potential extreme losses. We systematically adjust $\alpha$ values from $0.5$ to $1.0$ to observe changes in MAE for both VaR and CVaR settings. Then we further explore the model's sensitivity to different proportions of data considered under extreme risk conditions.

As shown in Fig.~\ref{fig:lambda_var_mse} and Fig.~\ref{fig:alpha_cvar_mse}, the results demonstrate a clear trend where increasing $\alpha$ values lead to a decrease in MAE, with an enhanced prediction accuracy with higher confidence levels. Specifically, the CVaR measure shows a more pronounced reduction in MAE than VaR, indicating its superior capability in managing tail risks.

For VaR, the MAE is higher at lower $\alpha$ levels, which implies that a lower proportion of data considered at risk leads to reduced prediction accuracy. Conversely, the accuracy improves as the $\alpha$ value increases, indicating a more conservative risk assessment. This trend underscores the importance of carefully selecting the $\alpha$ parameter to achieve an optimal balance between risk sensitivity and prediction accuracy in financial modeling, particularly in environments where accurate risk assessment is crucial. As CVaR focuses on averaging losses beyond the VaR threshold, it is more effective in scenarios demanding stringent risk management. Thus, it is a valuable tool for enhancing model reliability under extreme conditions.

\section{Conclusion and Future Work}

In conclusion, our paper proposes a new loss function, Loss-at-Risk, integrating risk assessment into Transformer models' fine-tuning process for financial forecasting. In Loss-at-Risk, we incorporate both VaR and CVaR with vanilla MSE loss to improve the models' capabilities to accurately assess and manage extreme financial risks. Our experiments with highly volatile financial datasets clearly demonstrate the enhanced ability of these Transformer models to predict and manage extreme losses. In this way, we enable safer and more reliable financial tools for decision-making processes.

For future work, further research could explore the impact of the Loss-at-Risk on large language models in various real-world scenarios to deepen the understanding of its practical applications and effectiveness in risk-sensitive environments. This could lead to a broader adoption of risk-aware methodologies across different sectors, and enhance the large language models' reliability and utility in complex decision-making scenarios.

\section*{Acknowledgment}

The authors Jinghan Zhang, Xinhao Zhang and Kunpeng Liu are supported by NSF 2348485 and NSF 2426339.


\bibliographystyle{ieeetr}
\bibliography{ref}

\vspace{12pt}
\end{document}